\documentclass[twoside,leqno,twocolumn]{article}
\usepackage{ltexpprt}
\usepackage{graphicx}
\usepackage{amsfonts}
\usepackage{booktabs}
\usepackage{url}

\begin{document}

\title{Spatial Variational Auto-Encoding via Matrix-Variate Normal Distributions}
\author{Zhengyang Wang\thanks{Department of Computer Science and Engineering at Texas A\&M University. Email: zhengyang.wang@tamu.edu} \\
\and
Hao Yuan\thanks{School of Electrical Engineering and Computer Science at Washington State University. Email: hao.yuan@wsu.edu} \\
\and
Shuiwang Ji\thanks{Department of Computer Science and Engineering at Texas A\&M University. Email: sji@tamu.edu}}
\date{}

\maketitle


\fancyfoot[R]{\scriptsize{Copyright \textcopyright\ 2019 by SIAM\\
Unauthorized reproduction of this article is prohibited}}





\begin{abstract} \small\baselineskip=9pt
The key idea of variational auto-encoders~(VAEs) resembles that of
traditional auto-encoder models in which spatial information is supposed to
be explicitly encoded in the latent space. However, the latent variables in
VAEs are vectors, which can be interpreted as multiple feature maps of size
1x1. Such representations can only convey spatial information implicitly when
coupled with powerful decoders. In this work, we propose spatial VAEs that
use feature maps of larger size as latent variables to explicitly capture
spatial information. This is achieved by allowing the latent variables to be
sampled from matrix-variate normal~(MVN) distributions whose parameters are
computed from the encoder network. To increase dependencies among locations
on latent feature maps and reduce the number of parameters, we further
propose spatial VAEs via low-rank MVN distributions. Experimental results
show that the proposed spatial VAEs outperform original VAEs in capturing
rich structural and spatial information.
\end{abstract}

\section*{Keywords}
Deep learning, variational auto-encoders, matrix-variate normal distributions, generative models, unsupervised learning

\section{Introduction.}\label{intro}
The mathematical and computational modeling of probability
distributions in high-dimensional space and generating samples from
them are highly useful yet very challenging. With the development of
deep learning methods, deep generative models have been shown to be
effective and
scalable~\cite{kingma2013auto,rezende2014stochastic,burda2015importance,gulrajani2016pixelvae,makhzani2015adversarial,goodfellow2014generative,radford2015unsupervised}
in capturing probability distributions over high-dimensional data
spaces and generating samples from them. Among them, variational
auto-encoders~(VAEs)~\cite{kingma2013auto,rezende2014stochastic,doersch2016tutorial,kingma2016improved} are one of the most
promising approaches. In machine learning, the auto-encoder
architecture is applied to train scalable models by learning latent
representations. For image modeling tasks, it is preferred to encode
spatial information into the latent space explicitly. However, the
latent variables in VAEs are vectors, which can be interpreted as
$1\times 1$ feature maps with no explicit spatial information. While
such lack of explicit spatial information does not lead to major
performance problems on simple tasks such as digit generation from
the MNIST dataset~\cite{lecun1998mnist}, it greatly limits the
model's abilities when images are more
complicated~\cite{krizhevsky2009learning,liu2015faceattributes}.

To overcome this limitation, we propose spatial VAEs that employ
$d\times d$ $(d>1)$ feature maps as latent representations. Such
latent feature maps are generated from matrix-variate normal~(MVN)
distributions whose parameters are computed from the encoder
network. Specifically, MVN distributions are able to generate
feature maps with appropriate dependencies among locations. To
increase dependencies among locations on latent feature maps and
reduce the number of parameters, we further propose spatial VAEs via
low-rank MVN distributions. In this low-rank formulation, the mean
matrix of MVN distribution is computed as the outer product of two
vectors computed from the encoder network. Experimental results on
image modeling tasks demonstrate the capabilities of our spatial
VAEs in complicated image generation tasks.

It is worth noting that the original VAEs can be considered as a
special case of spatial VAEs via MVN distributions. That is, if we
set the size of feature maps generated via MVN distributions to
$1\times 1$, spatial VAEs via MVN distributions reduce to the
original VAEs. More importantly, when the size of feature maps is
larger than $1\times 1$, direct structural ties have been built into
elements of the feature maps via MVN distributions. Thus, our
proposed spatial VAEs are intrinsically different with the original
VAEs when the size of feature maps is larger than $1\times 1$.
Specifically, our proposed spatial VAEs cannot be obtained by
enlarging the size of the latent representations in the original
VAEs.

\section{Background and Related Work.}\label{bg}

In this section, we introduce the architectures of auto-encoders and variational auto-encoders.

\subsection{Auto-Encoder Architectures.}\label{ae}

Auto-encoder (AE) is a model architecture used in tasks like image
segmentation~\cite{zeiler2010deconvolutional,ronneberger2015u,long2015fully},
machine translation~\cite{bahdanau2014neural,sutskever2014sequence}
and denoising reconstruction~\cite{vincent2008extracting,vincent2010stacked}. It
consists of two parts: an encoder that encodes the input data into
lower-dimensional latent representations and a decoder that
generates outputs by decoding the representations. Depending on
different tasks, the latent representations will focus on different
properties of input data. Nevertheless, these tasks usually require
outputs to have similar or exactly the same structure as inputs.
Thus, structural information is expected to be preserved through the
encoder-decoder process.

In computer vision tasks, structural information usually means
spatial information of images. There are two main strategies to
preserve spatial information in AE for image tasks. One is to apply
very powerful decoders, like conditional pixel convolutional neural
networks~(PixelCNNs)~\cite{oord2016pixel,van2016conditional,salimans2017pixelcnn++,gulrajani2016pixelvae},
that generate output images pixel-by-pixel. In this way, the
decoders can recover spatial information in the form of dependencies
among pixels. However, pixel-by-pixel generation is very slow,
resulting in major speed problems in practice. The other method is
to let the latent representations explicitly contain spatial
information and apply decoders that can make use of such
information. To apply this strategy for image tasks, usually the
latent representations are feature maps of size between the size of
a pixel~($1\times 1$) and that of the input image, while the
decoders are deconvolutional neural
networks~(DCNNs)~\cite{zeiler2010deconvolutional}.
Since most computer vision tasks only require high-level spatial
information like relative locations of objects instead of detailed
relationships among pixels, preserving only rough spatial
information is enough, and this strategy is proved effective and
efficient.

\subsection{Variational Auto-Encoders.}\label{vae}

In unsupervised learning, generative models aim to modeling the
underlying data distribution. Formally, for data
space~$\mathcal{X}$, let $p_{true}(x)$ denote the probability
density function (PDF) of the true data distribution for~$x\in
\mathcal{X}$. Given a dataset~$\mathcal{D}=\{x^{(i)}\}^N_{i=1}$ of
i.i.d samples from~$\mathcal{X}$, generative models try to
approximate $p_{true}(x)$ using a model distribution $p_\theta(x)$
where $\theta$ represents model parameters. To train the model,
maximum likelihood (ML) inference is performed on $\theta$; that is,
parameters are updated to optimize $\log p_\theta(\mathcal{D})=\log
p_\theta(x^{(1)},\dots,x^{(N)})=\sum_{i=1}^{N}\log
p_\theta(x^{(i)}).$ The approximation quality of $p_\theta(x)$
relies on the generalization ability of the model. In machine
learning, it highly depends on learning latent representations which
can encode common features among data samples and disentangle
abstract explanatory factors behind the
data~\cite{bengio2013representation}. In data generation tasks, we
apply $p_\theta(x)=\int p_\theta(x|z)p_\theta(z)dz$ for modeling,
where $p_\theta(z)$ is the PDF of the distribution of latent
representations and $p_\theta(x|z)$ represents a complex mapping
from the latent space to the data space. A major advantage of using
latent representations is dimensionality reduction of data since
they are low-dimensional. The prior $p_\theta(z)$ can be simple and
easy to model while the mapping represented by $p_\theta(x|z)$ can
be learned through complicated deep learning models automatically.

Recently, \cite{kingma2013auto} point out that the above model has
intractability problems and can only be trained by costly
sampling-based methods. To tackle this, they propose variational
auto-encoders~(VAEs), which instead maximize a variational lower
bound of the log-likelihood as
\begin{eqnarray}\label{eqn:2}
\log p_\theta(x)&\geq& \mathcal{L}_{\mbox{{\scriptsize{VAE}}}}\nonumber\\
&=&\mathbb{E}_{z\sim q_\phi(z|x)}[\log p_\theta(x|z)]\nonumber\\
&&-D_{KL}[q_\phi(z|x)|p_\theta(z)],
\end{eqnarray}
where $q_\phi(z|x)$ is an approximation model to the intractable
$p_\theta(z|x)$, parameterized by $\phi$, $D_{KL}[\cdot]$ represents
the Kullback-Leibler divergence. In VAEs,
$p_\theta(x|z)=\mathcal{N}(x;f_\theta(z),\sigma^2I)$,
$q_\phi(z|x)=\mathcal{N}(z;\mu_\phi(x),\Sigma_\phi(x))$, and
$p_\theta(z)=\mathcal{N}(z;0,I)$ are modeled as multivariate
Gaussian distributions with diagonal covariance matrices. Here,
$f_\theta(z)$, $\mu_\phi(x)$ and $\Sigma_\phi(x)$ are computed with
deep neural networks like CNNs. Figure~\ref{fig:f1} shows the
architecture of VAEs. The model parameters $\theta$ and $\phi$ can
be trained using the reparameterization
trick~\cite{rezende2014stochastic}, where the sampling process
$z\sim q_\phi(z|x)=\mathcal{N}(z;\mu_\phi(x),\Sigma_\phi(x))$ is
decomposed into two steps as
\begin{eqnarray}\label{eqn:3}
&&\epsilon \sim \mathcal{N}(\epsilon;0,I),\nonumber\\
&&z=\mu_\phi(x)+\Sigma_\phi^{\frac{1}{2}}(x)\ast \epsilon.
\end{eqnarray}

\section{Spatial Variational Auto-Encoders.}\label{main}

In this section, we analyze a problem of the original VAEs and
propose spatial VAEs in Section~\ref{svae} to overcome it.
Afterwards, several ways to implement spatial VAEs are discussed. A
na\"{\i}ve implementation is introduced and analyzed in
Section~\ref{naive}, followed by a method that incorporates the
use of matrix-variate normal~(MVN) distributions in
Section~\ref{mvnd}. Finally, we propose our final model, spatial
VAEs via low-rank MVN distributions, by applying a low-rank
formulation of MVN distributions in Section~\ref{trick}.

\subsection{Overview.}\label{svae}

Note that $p_\theta(x|z)$ and $q_\phi(z|x)$ in VAEs resemble the
encoder and decoder, respectively, in AE for image reconstruction
tasks, where $z$ represents the latent representations. However, in
VAE, $z$ is commonly a vector, which can be considered as multiple
$1\times 1$ feature maps. While $z$ may implicitly preserve some
spatial information of the input image~$x$, it raises the
requirement for a more complex decoder. Given a fixed architecture,
the hypothesis space of decoder models is limited. As a result, the
optimal decoder may not lie in the hypothesis
space~\cite{zhao2017towards}. This problem significantly hampers the
performance of VAEs, especially when spatial information is
important for images in~$\mathcal{X}$.

Based on the above analysis, it is beneficial to either have larger
hypothesis space for decoders or let $z$ explicitly contain spatial
information. Note that these two methods correspond to the two
strategies introduced in Section~\ref{ae}.
\cite{gulrajani2016pixelvae} follow the first strategy and propose
PixelVAEs whose decoders are conditional
PixelCNNs~\cite{van2016conditional} instead of simple DCNNs. As
conditional PixelCNNs themselves are also generative models,
PixelVAEs can be considered as conditional PixelCNNs with the
conditions replaced by $z$. In spite of their impressive results,
the performance of PixelVAEs and conditional PixelCNNs is similar,
which indicates that conditional PixelCNNs are responsible for
capturing most properties of images in~$\mathcal{X}$. In this case,
$z$ contributes little to the performance. In addition, applying
conditional PixelCNNs leads to very slow generation process in
practice. In this work, the second strategy is explored by
constructing spatial latent representations $z$ in the form of
feature maps of size larger than $1\times 1$. Such feature maps can
explicitly contain spatial information. We term VAEs with spatial
latent representations as spatial VAEs.

The main distinction between spatial VAEs and the original VAEs is
the size of latent feature maps. By having $d\times d$ $(d>1)$
feature maps instead of $1\times 1$ ones, the total dimension of the
latent representations $z$ significantly increases. However, spatial
VAEs are essentially different from the original VAEs with a
higher-dimensional latent vector $z$. Suppose the vector $z$ is
extended by $d^2$ times in order to match the total dimension, the
number of hidden nodes in each layer of decoders will explode
correspondingly. This results in an explosion in the number of
decoders' parameters, which slows down the generation process.
Whereas in spatial VAEs, decoders becomes even simpler since
$d\times d$ is closer to the required size of output images. From
the other side, when using decoders of similar capacities, spatial
VAEs must have higher-dimensional latent representations than the
original VAEs. It is demonstrated that this only slightly influences
the training process by requiring more outputs from encoders, while
the generation process that only involves decoders remains
unaffected. Our experimental results show that with proper designs,
spatial VAEs substantially outperform the original VAEs when
applying similar decoders.

\begin{figure*}[!t]
	\centering
	\includegraphics[width=\textwidth]{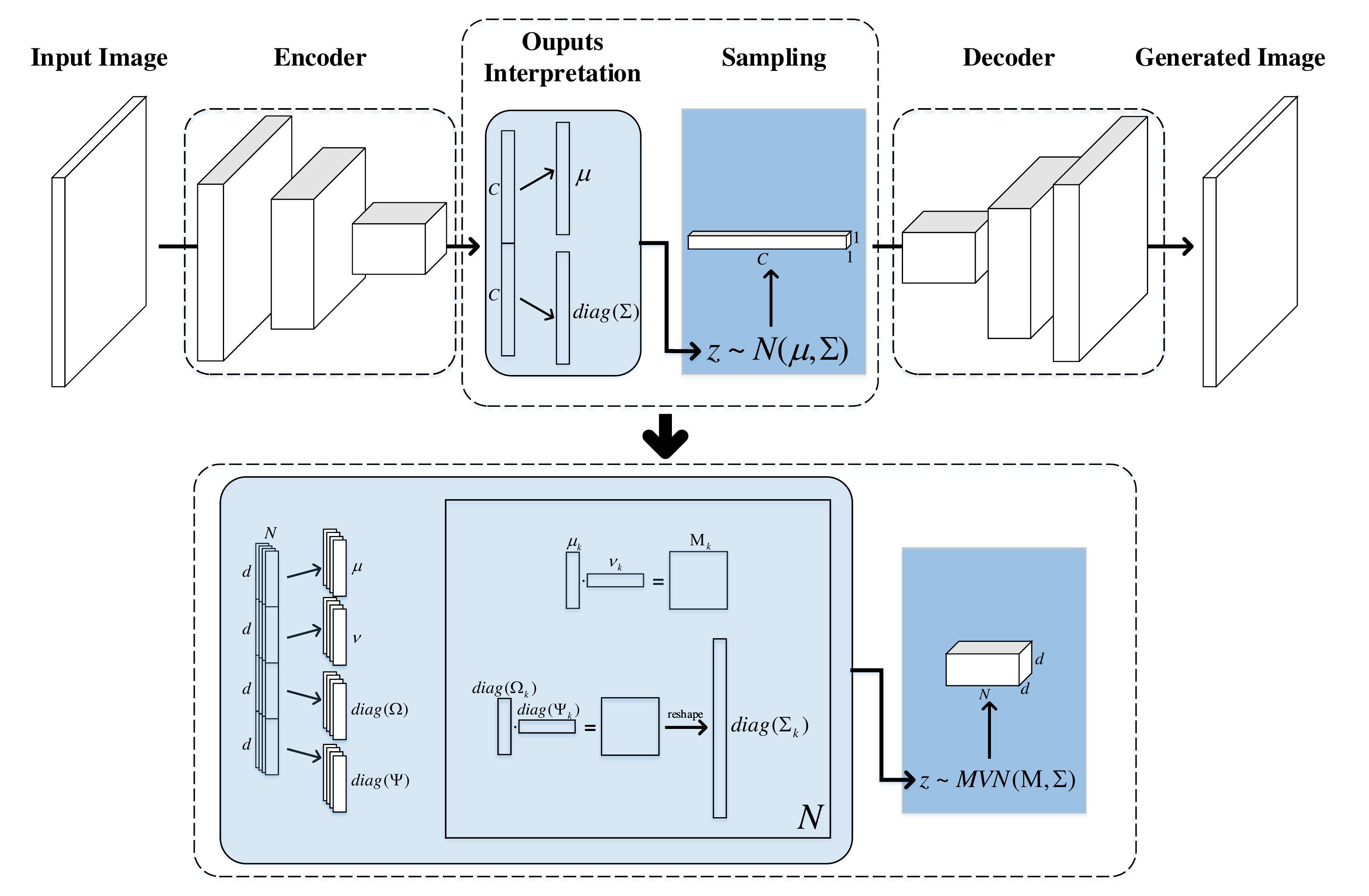}
	\caption[LoF entry]{Illustration of the differences between the
		proposed spatial VAEs via low-rank MVN distributions and the
		original VAEs. At the top is the architecture of the original VAEs
		where the latent $z$ is a vector sampled from a multivariate
		Gaussian distribution with a diagonal covariance matrix. Below is
		the proposed model which is explained in detail in
		Section~\ref{trick}. Briefly, it modifies the sampling process by
		incorporating a low-rank formulation of the MVN distributions and
		produces latent representations that explicitly retain spatial
		information.} \label{fig:f1}
\end{figure*}

\subsection{Na\"{\i}ve Spatial VAEs.}\label{naive}


To achieve spatial VAEs, a direct and na\"{\i}ve way is to simply
reshape the original vector $z$ into $N$ feature maps of size
$d\times d$. But this na\"{\i}ve way is problematic since the
sampling process does not change. Note that in the original VAEs,
the vector $z$ is sampled from
$q_\phi(z|x)=\mathcal{N}(z;\mu_\phi(x),\Sigma_\phi(x))$. The
covariance matrix $\Sigma_\phi(x)$ is diagonal, meaning each
variable is uncorrelated. In particular, for multivariate Gaussian
distributions, uncorrelation implies independence. Therefore, $z$'s
components are independent random variables and the variances of
their distributions correspond to entries on the diagonal of
$\Sigma_\phi(x)$. Specifically, suppose $z$ is a $C$-dimensional
vector, the $i^{th}$ component is a random variable that follows the
univariate normal distribution as
$z_i\sim\mathcal{N}(z_i;\mu_\phi(x)_i,\mbox{diag}(\Sigma_\phi(x))_i),\
i=1,\dots,C,$
where $\mbox{diag}(\cdot)$ represents the vector consisting of a
matrix's diagonal entries. After applying the reparameterization
trick, we can rewrite Equation~\ref{eqn:3} as
\begin{eqnarray}\label{eqn:5}
&&\epsilon_i \sim \mathcal{N}(\epsilon_i;0,1),\nonumber\\
&&z_i=\mu_\phi(x)_i+\mbox{diag}(\Sigma_\phi(x))_i^{\frac{1}{2}}\ast \epsilon_i,\ i=0,\dots,C.
\end{eqnarray}
To sample $N$ feature maps of size $d\times d$ in na\"{\i}ve spatial
VAEs, the above process is followed by a reshape operation while
setting $C=d^2N$.

However, between two different components $z_i$ and $z_j$, the only
relationship is that their respective distribution parameters
$(\mu_\phi(x)_i,\mbox{diag}(\Sigma_\phi(x))_i)$ and
$(\mu_\phi(x)_j,\mbox{diag}(\Sigma_\phi(x))_j)$ are both computed
from $x$. Such dependencies are implicit and weak. It is obvious
that after reshaping, there is no direct relationship among locations
within each feature map, while spatial latent representations should
contain spatial information like dependencies among locations. To
overcome this limitation, we propose spatial VAEs via matrix-variate
normal distributions.

\subsection{Spatial VAEs via Matrix-Variate Normal Distributions.}\label{mvnd}

Instead of obtaining $N$ feature maps of size $d\times d$ by first
sampling a $d^2N$-dimensional vector from multivariate normal
distributions and then reshaping, we propose to directly sample
$d\times d$ matrices as feature maps from matrix-variate
normal~(MVN) distributions~\cite{gupta1999matrix}, resulting in an
improved model known as spatial VAEs via MVN distributions.
Specifically, we modify $q_\phi(z|x)$ in the original VAEs and keep
other parts the same. As explained below, MVN distributions can
model dependencies between the rows and columns in a matrix. In this
way, dependencies among locations within a feature map are
established. We proceed by providing the definition of MVN
distributions.

\textbf{Definition:} A random matrix $A\in\mathbb{R}^{m\times n}$ is said
to follow a matrix-variate normal distribution
$\mathcal{N}_{m,n}(A;M,\Omega\otimes\Psi)$ with mean matrix
$M\in\mathbb{R}^{m\times n}$ and covariance matrix
$\Omega\otimes\Psi$, where $\Omega\in\mathbb{R}^{m\times m} > 0$, $\Psi\in\mathbb{R}^{n\times n} > 0$,
if $\mbox{vec}(A^T)$ follows the multivariate normal distribution
$\mathcal{N}(\mbox{vec}(A^T);\mbox{vec}(M^T),\Omega\otimes\Psi)$.
Here, $\otimes$ denotes the Kronecker product and
$\mbox{vec}(\cdot)$ denotes transforming a $\mathbb{R}^{m\times n}$
matrix into an $mn$-dimensional vector by concatenating the columns.

In MVN distributions, $\Omega$ and $\Psi$ capture the relationships
across rows and columns, respectively, of a matrix. By constructing
the covariance matrix through the Kronecker product of these two
matrices, dependencies among values in a matrix can be modeled. In
spatial VAEs, a feature map $F$ can be considered as a
$\mathbb{R}^{d\times d}$ matrix that follows a MVN distribution
$\mathcal{N}_{d,d}(F;M,\Omega\otimes\Psi)$, where $\Omega\in\mathbb{R}^{d\times d}$ and
$\Psi\in\mathbb{R}^{d\times d}$ are diagonal matrices. Although within $F$
the random variables corresponding to each location are still
independent since $\Omega\otimes\Psi$ is diagonal, MVN distributions
are able to add direct structural ties among locations through their
variances. For example, for two locations $(i_1,j_1)$ and
$(i_2,j_2)$ in $F$,
\begin{eqnarray}\label{eqn:6}
&&F_{(i_1,j_1)}\sim\mathcal{N}(F_{(i_1,j_1)};M_{(i_1,j_1)},\mbox{diag}(\Omega\otimes\Psi)_{i_1\ast j_1}),\\
&&F_{(i_2,j_2)}\sim\mathcal{N}(F_{(i_2,j_2)};M_{(i_2,j_2)},\mbox{diag}(\Omega\otimes\Psi)_{i_2\ast j_2}).
\end{eqnarray}
Here, $F_{(i_1,j_1)}$ and $F_{(i_2,j_2)}$ are independently sampled
from two univariate Gaussian distributions. However, the variances
$\mbox{diag}(\Omega\otimes\Psi)_{i_1\ast j_1}$ and
$\mbox{diag}(\Omega\otimes\Psi)_{i_2\ast j_2}$ have built direct
interactions through the Kronecker product. Based on this, we
propose spatial VAEs via MVN distributions, which samples $N$
feature maps of size $d\times d$ from $N$ independent MVN
distributions as
\begin{eqnarray}\label{eqn:7}
&&F_k\sim\mathcal{N}_{d,d}(F_k;{M_k}_\phi(x),{\Omega_k}_\phi(x)\otimes{\Psi_k}_\phi(x)),\\
&&k=0,\dots,N,\nonumber
\end{eqnarray}
where ${M_k}_\phi(x)$, ${\Omega_k}_\phi(x)$ and ${\Psi_k}_\phi(x)$
are computed through the encoder. Here, compared to the original
VAEs, $q_\phi(z|x)$ is replaced but $p_\theta(z)$ remains the same.
Since MVN distributions are defined based on multivariate Gaussian
distributions, the term $D_{KL}[q_\phi(z|x)|p_\theta(z)]$ in
Equation~\ref{eqn:2} can be calculated in a similar way.

To demonstrate the differences with na\"{\i}ve spatial VAEs, we
reexamine the original VAEs. Note that na\"{\i}ve spatial VAEs have
the same sampling process as the original VAEs. The original VAE
samples a $C=d^2N$-dimensional vector $z$ from
$q_\phi(z|x)=\mathcal{N}(z;\mu_\phi(x),\Sigma_\phi(x))$ where $\mu_\phi(x)$ is a
$C$-dimensional vector and $\Sigma_\phi(x)$ is a
$\mathbb{R}^{C\times C}$ diagonal matrix. Because $\Sigma_\phi(x)$
is diagonal, it can be represented by the $C$-dimensional vector
$\mbox{diag}(\Sigma_\phi(x))$. To summarize, the encoder of the
original VAEs outputs $2C=2d^2N$ values which are interpreted as
$\mu_\phi(x)$ and $\mbox{diag}(\Sigma_\phi(x))$.

In spatial VAEs via MVN distributions, according to
Equation~\ref{eqn:7}, ${M_k}_\phi(x)$ is a $\mathbb{R}^{d\times d}$
matrix while ${\Omega_k}_\phi(x)$ and ${\Psi_k}_\phi(x)$ are
$\mathbb{R}^{d\times d}$ diagonal matrices that can be represented
by $d$-dimensional vectors. In this case, the required number of
outputs from the encoder is changed to $(d^2+2d)N$, corresponding to
$[{M_1}_\phi(x),\dots,{M_N}_\phi(x)]$,
$[\mbox{diag}({\Omega_1}_\phi(x)),\dots,\mbox{diag}({\Omega_N}_\phi(x))]$
and
$[\mbox{diag}({\Psi_1}_\phi(x)),\dots,\mbox{diag}({\Psi_N}_\phi(x))]$.
As has been explained in Section~\ref{naive}, since
${\Omega_k}_\phi(x)\otimes{\Psi_k}_\phi(x)$ is diagonal, sampling
the matrix $F_k$ is equivalent to sampling $d\times d$ scalar
numbers from $d\times d$ independent univariate normal
distributions. So the modified sampling process with the
reparameterization trick is
\begin{eqnarray}\label{eqn:8}
&&\epsilon_{(i,j,k)}\sim \mathcal{N}(\epsilon_{(i,j,k)};0,1),\nonumber\\
&&z_{(i,j,k)}={{\mu_k}_\phi(x)}_{(i,j)}\nonumber\\
&&\quad\quad\quad\quad+\mbox{diag}({\Omega_k}_\phi(x)\otimes{\Psi_k}_\phi(x))_{i\ast j}^{\frac{1}{2}}\ast \epsilon_{(i,j,k)},\\
&&i,j=0,\dots,d,\ k=1,\dots,N,\nonumber
\end{eqnarray}
where
\begin{eqnarray}
&&\mbox{diag}({\Omega_k}_\phi(x)\otimes{\Psi_k}_\phi(x))_{i\ast j}\nonumber\\
&=&[\mbox{diag}({\Omega_k}_\phi(x))\mbox{diag}^T({\Psi_k}_\phi(x))]_{(i,j)}.\nonumber
\end{eqnarray}
Here, we take advantage of the fact that for diagonal matrices, the
Kronecker product is equivalent to the out-product of vectors. To be
specific, suppose $D_1$ and $D_2$ are two $\mathbb{R}^{d\times d}$
diagonal matrices, then $d_1=\mbox{diag}(D_1)$ and
$d_2=\mbox{diag}(D_2)$ are two $d$-dimensional vectors and satisfy
\begin{equation}\label{eqn:9}
\mbox{diag}(D_1\otimes D_2)=\mbox{vec}(d_1 d^T_2).
\end{equation}
It is worth noting that, compared to na\"{\i}ve spatial VAEs, the
required number of outputs from the encoder decreases from $2d^2N$
to $(d^2+2d)N$. As a result, spatial VAEs via MVN distributions
leads to a simpler model while adding structural ties among
locations. Note that the original VAEs can be considered as a special case of the
spatial VAEs via MVN distributions. That is, if we set $d=1$, spatial
VAEs via MVN distributions reduce to the original VAEs.

\begin{figure*}[!t]
	\centering
	\includegraphics[width=0.95\textwidth]{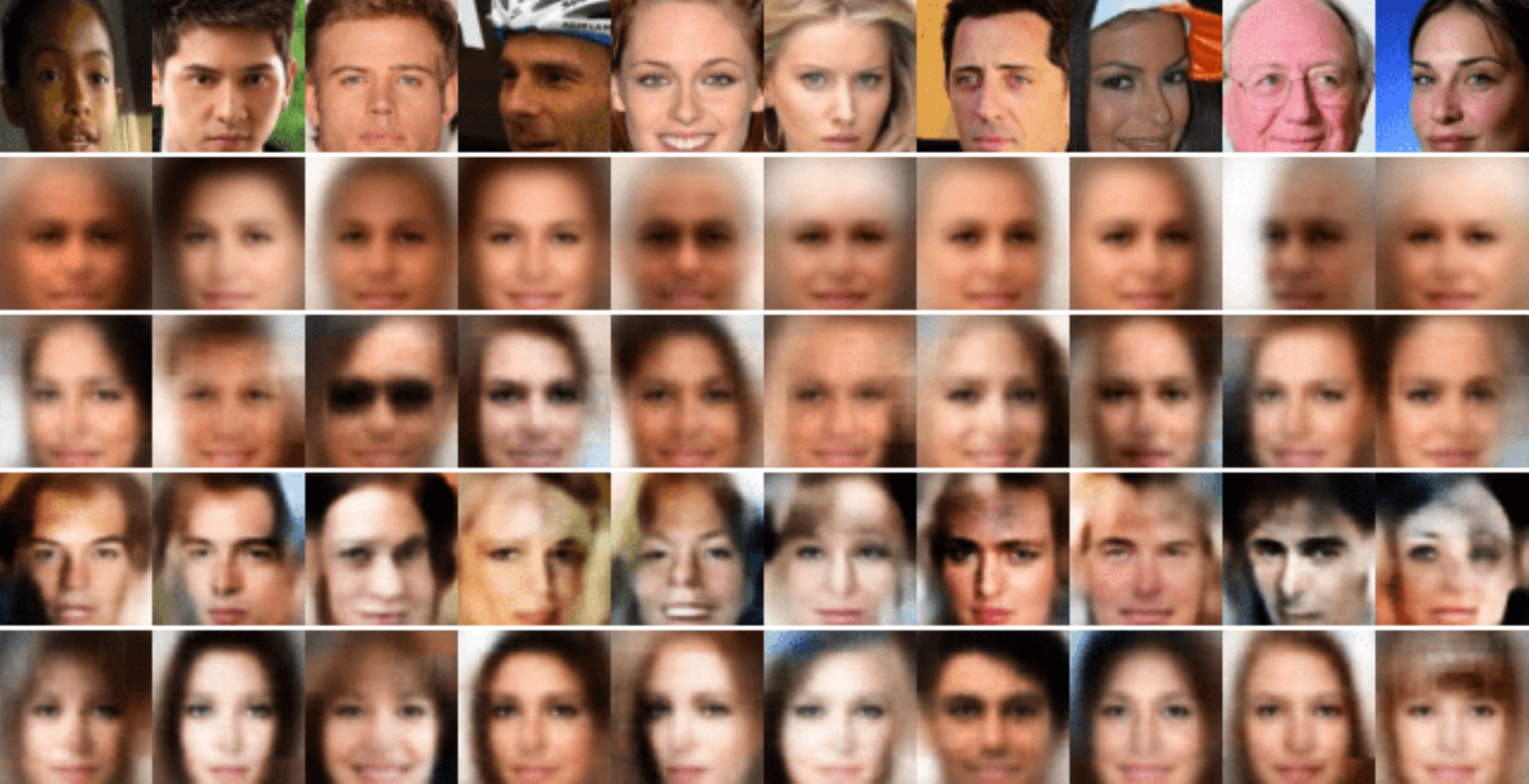}
	\caption[LoF entry]{Sample face images generated by different VAEs when trained on the CelebA dataset. The first and second rows shows training images and images generated by the original VAEs. The remaining three rows are the results of na\"{\i}ve spatial VAEs, spatial VAEs via MVN distributions and spatial VAEs via low-rank MVN distributions, respectively.}
	\label{fig:f2}
\end{figure*}

\subsection{A Low-Rank Formulation.}\label{trick}

The use of MVN distributions makes locations directly related to
each other within a feature map by adding restrictions on variances.
However, in probability theory, variance only measures the expected
distance from the mean. To have more direct relationships, it is
preferred to have restricted means. In this section, we introduce a
low-rank formulation of MVN
distributions~\cite{allen2010transposable} for spatial VAEs.

The low-rank formulation of a MVN distribution
$\mathcal{N}_{m,n}(M,\Omega\otimes\Psi)$ is denoted as
$\mathcal{N}_{m,n}(\mu,\nu,\Omega\otimes\Psi)$ where the mean matrix $M$ is
computed by the out-product $\mu\nu^T$ instead. Here, $\mu$ and $\nu$ are
$m$-dimensional and $n$-dimensional vectors, respectively. Similar to
computing the covariance matrix through the Kronecker product of two separate
matrices, it explicitly forces structural interactions among entries of the
mean matrix. Applying this low-rank formulation leads to our final model,
spatial VAEs via low-rank MVN distributions, which is illustrated in
Figure~\ref{fig:f1}. By using two distinct $d$-dimensional vectors to
construct ${M_i}_\phi(x)\in\mathbb{R}^{d\times d}$,
Equation~\ref{eqn:7} is modified as
\begin{eqnarray}\label{eqn:10}
&&F_k\sim\mathcal{N}_{d,d}(F_k;{\mu_k}_\phi(x){\nu_k}^T_\phi(x),{\Omega_k}_\phi(x)\otimes{\Psi_k}_\phi(x)),\\
&&\forall k=0,\dots,N,\nonumber
\end{eqnarray}
where ${\mu_k}_\phi(x)$ and ${\nu_k}_\phi(x)$ are $d$-dimensional
vectors. For the encoder, the number of outputs is further reduced
to $4dN$ from $(d^2+2d)N$, replacing $d^2N$ outputs for
$({M_1}_\phi(x),\dots,{M_N}_\phi(x))$ with $dN$ outputs for
$({\mu_1}_\phi(x),\dots,{\mu_N}_\phi(x))$ and another $dN$ outputs
for $({\nu_1}_\phi(x),\dots,{\nu_N}_\phi(x))$. In contrast to
Equation~\ref{eqn:8}, the two-step sampling process can be expressed
as
\begin{eqnarray}\label{eqn:11}
&&\epsilon_{(i,j,k)}\sim \mathcal{N}(\epsilon_{(i,j,k)};0,1),\nonumber\\
&&z_{(i,j,k)}=({{\mu_k}_\phi(x){\nu_k}^T_\phi(x)})_{(i,j)}\nonumber\\
&&\quad\quad\quad\quad+\mbox{diag}({\Omega_k}_\phi(x)\otimes{\Psi_k}_\phi(x))_{i\ast j}^{\frac{1}{2}}\ast \epsilon_{(i,j,k)},\\
&&i,j=0,\dots,d,\ k=1,\dots,N,\nonumber
\end{eqnarray}
where
\begin{eqnarray}
&&\mbox{diag}({\Omega_k}_\phi(x)\otimes{\Psi_k}_\phi(x))_{i\ast j}\nonumber\\
&=&[\mbox{diag}({\Omega_k}_\phi(x))\mbox{diag}^T({\Psi_k}_\phi(x))]_{(i,j)}.\nonumber
\end{eqnarray}
As has been demonstrated in Section~\ref{svae}, spatial VAEs require
more outputs from encoders than the original VAEs, which slows down
the training process. Spatial VAEs via low-rank MVN distributions
properly address the problem while achieving appropriate spatial
latent representations. According to the experimental results, they
outperform the original VAEs in several image generation tasks when
similar decoders are used.

\section{Experimental Studies.}\label{exp}



We use the original VAEs as the baseline models in our experiments, as most
recent improvements on VAEs are derived from the vector latent
representations and can be easily incorporated into our matrix-based models.
To elucidate the performance differences of various spatial VAEs, we compare
the results of three different spatial VAEs as introduced in
Section~\ref{main}; namely na\"{\i}ve spatial VAEs, spatial VAEs via MVN
distributions and spatial VAEs via low-rank MVN distributions. We train the
models on the CelebA, CIFAR-10 and MNIST datasets, and analyze sample
images generated from the models to evaluate the performance. For the same
task, the encoders of all compared models are composed of the same
convolutional neural networks~(CNNs) and a fully-connected output
layer~\cite{lecun1998gradient,krizhevsky2012imagenet}. While the
fully-connected layer may differ as required by different numbers of output
units, it only slightly affects the training process. As discussed in
Section~\ref{svae}, it is reasonable to compare spatial VAEs with the
original VAEs in the case that their decoders have similar architectures and
model capabilities. Therefore, following the original VAEs, deconvolutional
neural networks~(DCNNs) are used as decoders in spatial VAEs. Meanwhile, the
total number of trainable parameters in the decoders of all compared models
are set to be as similar as possible while accommodating different input
sizes.

\subsection{CelebA.}

\begin{figure*}[!t]
	\centering
	\includegraphics[width=0.95\textwidth]{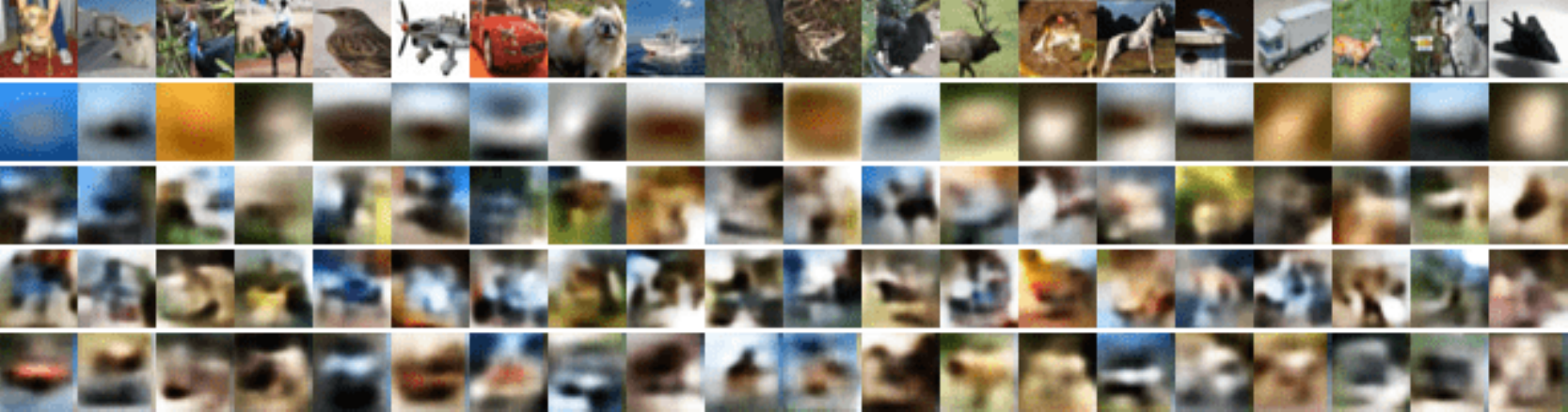}
	\caption[LoF entry]{Sample images generated by different VAEs when trained on the CIFAR-$10$ dataset. From top to bottom, the five rows are training images and images generated by the original VAEs, na\"{\i}ve spatial VAEs, spatial VAEs via MVN distributions, spatial VAEs via low-rank MVN distributions, respectively.}
	\label{fig:f3}
\end{figure*}

The CelebA dataset contains $202,599$ colored face images of size
$64\times 64$. The generative models are supposed to generate faces
that are similar but not exactly the same to those in the dataset.
For this task, the CNNs in the encoders have $3$ layers while the
decoders are $5$ or $6$-layer DCNNs corresponding to spatial VAEs
and the original VAEs, respectively. This difference is caused by
the fact that spatial VAEs have $d\times d$ ($d>1$) feature maps as
latent representations, which require fewer up-sampling operations
to obtain $64\times 64$ outputs. We set $d=3$ and $N=64$, and the
dimension of $z$ in the original VAEs is $81$ in order to have
decoders with similar numbers of trainable parameters.

Figure~\ref{fig:f2} shows sample face images generated by the
original VAEs and three different variants of spatial VAEs. It is
clear that spatial VAEs can generate images with more details than
the original VAEs. Due to the lack of explicit spatial information,
the original VAEs produce face images with little details like hair
near the borders. While na\"{\i}ve spatial VAEs seem to address this
problem, most faces have only incomplete hairs as na\"{\i}ve spatial
VAEs cannot capture the relationships among different locations.
Theoretically, spatial VAEs via MVN distributions are able to
incorporate interactions among locations. However, the results are
strange faces with some distortions. We believe the reason is that
adding dependencies among locations through restrictions on
distribution variances is not effective and sufficient. Spatial VAEs via
low-rank MVN distributions that have restricted means tackle this well and
generate faces with appealing visual appearances.

\subsection{CIFAR-10.}

The CIFAR-$10$ dataset consists of $60,000$ color images of $32\times 32$ in
$10$ classes. VAEs usually perform poorly in generating photo-realistic
images since there are significant differences among images in different
classes, indicating that the underlying true distribution of the data is a
multi-model. In this case, VAEs tend to output very blurry
images~\cite{theis2015note,goodfellow2014generative,goodfellow2016nips}.
However, comparison among different models can still demonstrate the
differences in terms of generative capabilities. In this experiment, we set
$d=3$ and $N=128$, and the dimension of $z$ in the original VAEs is $150$.
The encoders have $4$ layers while the decoders have $4$ or $5$
layers.

Some sample images are provided in Figure~\ref{fig:f3}. The original
VAEs only produce images composed of several colored areas, which is
consistent to the results of a similar model reported
in~\cite{rezende2014stochastic}. It is obvious that all three
implementations of spatial VAEs generate images with more details.
However, na\"{\i}ve spatial VAEs still produce meaningless images as
there is no relationship among different parts. The images generated
by spatial VAEs via MVN distributions look like some distorted
objects, which have similar problems to the results of the CelebA
dataset. Again, spatial VAEs via low-rank MVN distributions
outperform the other models, producing blurry but object-like
images.

\begin{table}[!t]
	\centering \caption{Parzen window log-likelihood estimates of test
		data on the MNIST dataset. We follow the same procedure as
		in~\cite{goodfellow2014generative}.} \label{table:1}
	\begin{tabular}{ l | c }
		\hline
		\textbf{Model}                               & \textbf{Log-Likelihood}  \\ \hline
		Original VAE                                 &  297  \\
		Na\"{\i}ve SVAE                              &  275  \\
		SVAE via MVN                                 &  267  \\
		SVAE via low-rank MVN                        &  296  \\
		\hline
	\end{tabular}
\end{table}

\subsection{MNIST.}

\begin{table*}[!t]
	\centering
	\caption{Training and generation time of different models when trained on the CelebA dataset using a Nvidia Tesla K40C GPU. The average time for training one epoch and the time for generating $10,000$ images are reported and compared.}
	\label{table:2}
	\begin{tabular}{ l | c | c }
		\hline
		\textbf{Model}                               & \textbf{Training time} & \textbf{Generation time} \\ \hline
		Original VAE                                 &  167.0309s             &  1.3892s \\
		Na\"{\i}ve SVAE                              &  178.8601s             &  1.3676s \\
		SVAE via MVN                                 &  177.4387s             &  1.3767s \\
		SVAE via low-rank MVN                        &  172.9639s             &  1.3686s \\
		\hline
	\end{tabular}
\end{table*}

We perform quantitative analysis on real-valued MNIST dataset by employing
the Parzen window log-likelihood estimates~\cite{breuleux2011quickly}. This
evaluation method is used for several generative models where the exact
likelihood is not
tractable~\cite{goodfellow2014generative,makhzani2015adversarial}. The
results are reported in Table~\ref{table:1} where SVAE is short for spatial
VAE. Despite of the difference in visual quality of generated images, spatial
VAE via low-rank MVN distributions shares similar quantitative results with
the original VAE. Note that generative models for images are supposed to
capture the underlying data distribution by maximizing log-likelihood and
generate images that are similar to real ones. However, it has been pointed
in \cite{theis2015note} that these two objectives are not consistent, and
generative models need to be evaluated directly with respect to the
applications for which they were intended. A model that can generates samples
with good visual appearances may have poor average log-likelihood on test
dataset and vice versa. Common examples of deep generative models are VAEs
and generative adversarial networks~(GANs)~\cite{goodfellow2014generative}.
VAEs usually have higher average log-likelihood while GANs can generate more
photo-realistic images. This is basically caused by the different training
objectives of these two models~\cite{goodfellow2016nips}. Currently there is
no commonly accepted standard for evaluating
generative models.

\subsection{Timing Comparison.}

To show the influence of different spatial VAEs to the training
process, we compare the training time on the CelebA dataset.
Theoretically, spatial VAEs slow down training due to the larger
numbers of outputs from encoders. To keep the number of
trainable parameters in decoders roughly equal, we set the dimension
of $z$ in the original VAEs to be $81$ while $d=3$ and $N=64$ for
spatial VAEs. According to Section~\ref{main}, the numbers of
outputs from their encoders are $162$, $1152$, $960$, and $768$ for
the original VAE, na\"{\i}ve spatial VAE, spatial VAE via MVN
distributions and spatial VAE via low-rank MVN distributions,
respectively. We train our models on a Nvidia Tesla K40C GPU and
report the average time for training one epoch in
Table~\ref{table:2}. Comparisons of the time for generating $10,000$
images are also provided to show that the increase in the total
dimension of latent representations does not affect the generation
process.

The results show consistent relationships between the training time
and the number of outputs from encoders; that is, spatial VAEs cost
more time than the original VAE but spatial VAEs via low-rank MVN
distributions can alleviate this problem. Moreover, spatial VAEs
only slightly slow down the training process since they only affect
one single layer in the models.

\section{Conclusion.}\label{con}

In this work, we propose spatial VAEs for image generation tasks,
which improve VAEs by requiring the latent representations to
explicitly contain spatial information of images. Specifically, in
spatial VAEs, $d\times d$ ($d>1$) feature maps are sampled to serve
as spatial latent representations in contrast to a vector. This is
achieved by sampling the latent feature maps from MVN distributions,
which can model dependencies between the rows and columns in a
matrix. We further propose to employ a low-rank formulation of MVN
distributions to establish stronger dependencies. Qualitative
results on different datasets show that spatial VAEs via low-rank
MVN distributions substantially outperform the original VAEs.

\section*{Acknowledgements.}
This work was supported by the National Science Foundation grants IIS-1633359
and DBI-1641223.

\bibliographystyle{siamplain}
\bibliography{reference_vae}

\end{document}